(RESEARCH ARTICLE)

Check for updates

# FROM PATTERNS TO THREATS: A DEEP LEARNING MODEL FOR ANOMALY DETECTION IN CYBER-PHYSICAL SYSTEMS

Kayode L. Ogunsusi *

Department of Mathematics and Statistics, Austin Peay State University, Clarksville, TN 37044, USA.
* Corresponding Author

ORCID Details
Kayode L. Ogunsusi: https://orcid.org/0009-0002-7392-7141

## ABSTRACT

Cyber-physical systems increasingly rely on interconnected sensing, communication, and control components, creating a need for reliable detection of abnormal behavior. This study proposes a Deep Autoencoder for anomaly detection using the Edge-IIoTset dataset. After preprocessing the full dataset, 2,219,201 observations and 38 numerical features were retained. The proposed model employed a 38–24–12–6–12–24–38 encoder-decoder architecture and was trained exclusively on normal observations to learn representative patterns of expected system behavior. Anomaly decisions were based on reconstruction error, with the classification threshold selected independently from validation data by maximizing the F1-score. To assess model stability, five independent runs were conducted using different random seeds, and results were reported as mean ± standard deviation. The proposed Deep Autoencoder achieved an accuracy of 0.9195 ± 0.0061, precision of 0.9602 ± 0.0204, recall of 0.7348 ± 0.0095, F1-score of 0.8325 ± 0.0121, ROC-AUC of 0.8825 ± 0.0339, and PR-AUC of 0.8640 ± 0.0286. Compared with a Simple Autoencoder baseline, the proposed model improved recall and F1-score while reducing the false positive rate. The results demonstrate that the proposed reconstruction-based representation learning approach can provide an effective and reproducible method for detecting anomalous behavior in cyber-physical environments.

## 1 INTRODUCTION

Cyber-physical systems (CPS) integrate computational, communication, and physical processes across environments such as industrial control, smart infrastructure, transportation, and connected devices. Their increasing connectivity improves automation and operational efficiency but also expands the attack surface through which malicious or abnormal activity can affect both digital and physical processes [1]. Prior studies have therefore emphasized the importance of detecting deviations in system and sensor behavior before they develop into significant operational or security events [2].

Anomaly detection provides an important approach to this problem because it focuses on identifying observations that differ from expected system behavior rather than relying solely on predefined attack signatures. Traditional detection approaches can be limited when system relationships are highly nonlinear or when anomalous patterns are insufficiently represented during model development. Deep learning offers an alternative by learning complex representations directly from multidimensional observations, and deep anomaly-detection methods have increasingly been applied where normal and abnormal patterns cannot be adequately described using simple statistical boundaries [3].

Among deep-learning methods, autoencoders are particularly suitable for reconstruction-based anomaly detection because they can be trained primarily on normal observations. The model learns a compressed representation of expected behavior and attempts to reconstruct each input from this representation. Observations that differ substantially from the learned normal patterns generally produce higher reconstruction errors and can therefore be identified as potential anomalies. This approach is attractive for CPS environments in which attack types may evolve and exhaustive labeling of abnormal conditions may be difficult.

This study develops a **Deep Autoencoder for anomaly detection in cyber-physical systems** using the Edge-IIoTset dataset, a realistic IoT/IIoT cybersecurity dataset introduced in 2022 with normal traffic and fourteen attack scenarios [4]. Unlike a conventional supervised attack classifier, the proposed model is trained using normal behavior and applies validation-based reconstruction-error thresholding to identify anomalous observations. Its performance is evaluated across five independent runs and compared with a simpler Autoencoder baseline.

The main contributions of this study are:

- Development of a 38–24–12–6–12–24–38 Deep Autoencoder for reconstruction-based anomaly detection.
- A validation-driven threshold-selection strategy that avoids using the test set to determine the anomaly decision boundary.
- A five-run experimental evaluation demonstrating improved recall, F1-score, ROC-AUC, and PR-AUC, together with a lower false positive rate, relative to a Simple Autoencoder baseline.

The remainder of the paper reviews related work, describes the dataset and preprocessing procedure, presents the proposed model and experimental design, reports the results, and discusses the findings and limitations.

## 2 RELATED WORK

Anomaly detection has long been used to identify deviations from expected network and system behavior, with earlier approaches relying on statistical analysis, clustering, support vector machines, and other conventional machine-learning techniques [5]. Although these methods have demonstrated effectiveness in intrusion detection, their performance can depend heavily on feature engineering and their ability to represent increasingly complex system behavior. In cyber-physical environments, this challenge is amplified by interactions among computational, communication, and physical components [1], [2].

The emergence of deep learning introduced methods capable of automatically learning nonlinear representations from high-dimensional data. Shone et al. [6] demonstrated the applicability of deep representation learning to network intrusion detection, illustrating how learned feature representations can support discrimination between normal and malicious activity. More broadly, deep anomaly-detection research has shown that neural architectures can capture complex patterns that may be difficult to characterize using conventional decision boundaries [8].

Autoencoders are particularly relevant to anomaly detection because they can learn compact representations of normal observations through reconstruction. Mirsky et al. [7], for example, introduced Kitsune, an online intrusion-detection approach based on an ensemble of autoencoders trained to distinguish normal network behavior from anomalous activity. Their findings demonstrated the practical potential of reconstruction-based neural models for identifying previously unseen network attacks without requiring conventional supervised attack classification.

Subsequent reviews have highlighted reconstruction-based learning as an important category of deep anomaly detection, particularly where anomalous examples are rare, diverse, or incompletely labeled [8]. In such settings, learning the characteristics of normal behavior and measuring deviations through reconstruction error provides an alternative to models that require comprehensive representations of individual attack categories.

For IoT and Industrial IoT environments, Ferrag et al. [4] introduced **Edge-IIoTset** in 2022 as a realistic cybersecurity dataset containing normal traffic and multiple attack scenarios across heterogeneous IoT and IIoT devices. The authors evaluated both conventional machine-learning and deep-learning approaches, establishing the dataset as a suitable benchmark for intrusion-detection research.

Building on these studies, the present work focuses specifically on reconstruction-based anomaly detection using a Deep Autoencoder trained on normal Edge-IIoTset observations. In addition to comparing the proposed architecture with a simpler Autoencoder baseline, the study evaluates performance over five independent runs and determines the anomaly threshold exclusively from validation data. This design enables assessment of not only predictive performance but also model stability and false-positive behavior.

## 3 DATASET AND PREPROCESSING

This section describes the dataset selection and preprocessing procedures used to prepare the data for deep-learning-based anomaly detection. The Edge-IIoTset dataset was cleaned to remove target leakage, non-informative variables, and unsuitable non-numeric attributes, after which the retained numerical features were partitioned into training, validation, and test sets. Missing-value treatment and feature standardization were performed using statistics derived only from normal training observations, and the final training data were restricted to normal behavior so that the autoencoder could learn representative system patterns and identify anomalous observations through reconstruction error.

### 3.1 Dataset Description

The experiments in this study were conducted using the Edge-IIoTset dataset, a publicly available cybersecurity dataset developed for evaluating machine-learning and deep-learning intrusion-detection methods in Internet of Things (IoT) and Industrial Internet of Things (IIoT) environments. The dataset was generated from a purpose-built testbed incorporating multiple sensors, IoT/IIoT devices, network protocols, edge-computing components, and industrial communication technologies. The original Edge-IIoTset testbed contains more than ten types of IoT devices and incorporates fourteen attack scenarios grouped into five broad threat categories, including denial-of-service/distributed denial-of-service, information-gathering, man-in-the-middle, injection, and malware attacks [4].

The DNN-EdgeIIoT-dataset.csv subset was used because it was specifically prepared by the dataset authors for evaluating deep-learning-based intrusion-detection systems. The selected file was obtained from the public Edge-IIoTset repository and processed as a binary anomaly-detection problem, where Attack_label = 0 represents normal behavior and Attack_label = 1 represents anomalous or attack behavior. The dataset authors separately provide Attack_type to identify specific attack categories; however, this variable was excluded from the predictor set in the present study to prevent direct target leakage [4].

After loading the complete DNN dataset**, 2,219,201 observations and 63 columns** were available. Following removal of the target variables, non-numeric attributes, and constant predictors, **38 numerical features** were retained for model development. The resulting binary class distribution consisted of approximately **72.8% normal observations and 27.2% anomalous observations.**

### 3.2 Feature Screening and Leakage Prevention

Feature preparation was performed before model training to ensure that the anomaly detector learned behavioral patterns rather than identifiers or variables that directly revealed the class label. Both Attack_label and Attack_type were removed from the predictor matrix. Attack_label served exclusively as the binary evaluation target, while Attack_type was excluded because knowledge of the attack category would provide direct information about whether an observation represented malicious behavior.

Non-numeric variables, including timestamps, source and destination host identifiers, HTTP request information, payload-related fields, MQTT strings, and other categorical network attributes, were also excluded from the present analysis. This decision produced a compact numerical feature space suitable for reconstruction-based deep anomaly detection while avoiding unnecessary expansion through categorical encoding. The original Edge-IIoTset preprocessing framework similarly identifies a set of address, payload, timestamp, and related fields for exclusion before model development.

Four numerical variables — *icmp.unused, http.tls_port, dns.qry.type,* and *mqtt.msg_decoded_as* were found to be constant throughout the analyzed dataset and were therefore removed because they contained no discriminative information. Infinite numerical values were converted to missing values before subsequent imputation. After these operations, the final predictor matrix contained **38 numerical features.**

### 3.3 Training, Validation, and Test Partitioning

The complete dataset was divided into mutually exclusive training, validation, and test subsets using a 70:15:15 stratified split. Stratification was applied using the binary attack label so that the proportion of normal and anomalous observations remained approximately constant across all partitions.

**Table 1: Final dataset partitioning**

| Dataset partition | Observations | Percentage | Normal | Anomalous |
|---|---|---|---|---|
| Training | 1,553,440 | 70% | 72.8% | 27.2% |
| Validation | 332,880 | 15% | 72.8% | 27.2% |
| Test | 332,881 | 15% | 72.8% | 27.2% |
| Total | **2,219,201** | **100%** | **72.8%** | **27.2%** |

### 3.4 Missing-Value Treatment and Feature Standardization

To preserve the one-class reconstruction-based training design, preprocessing statistics were estimated exclusively from normal observations in the training partition. This was an important design choice because allowing anomalous observations to influence feature imputation or scaling could partially incorporate characteristics of attack behavior into the representation learned as normal.

Missing numerical values were replaced using the **median value calculated from normal training observations**. Median imputation was selected because it is less sensitive than the arithmetic mean to extreme observations and skewed feature distributions.

Following imputation, each feature was standardized using:

$$z_{ij} = \frac{x_{ij} - \mu_j}{\sigma_j},$$

where $x_{ij}$ represents the value of feature *j* for observation *i*, while $\mu_j$ and $\sigma_j$ represent the mean and standard deviation of feature *j*, respectively, estimated exclusively from the normal training data.

The fitted imputation and standardization transformations were subsequently applied unchanged to the training, validation, and test partitions. All resulting feature matrices were converted to 32-bit floating-point representation to improve computational efficiency during neural-network training. No missing values remained after preprocessing.

### 3.5 Normal-Behavior Training Set

Unlike a conventional supervised binary classifier, the proposed autoencoder was trained **only on observations labeled as normal.** Of the 1,553,440 observations in the training partition, **1,130,950 normal observations** were used for model optimization. Similarly, **242,346 normal validation observations** were available for monitoring reconstruction performance during training.

The underlying premise is that an autoencoder trained exclusively on normal system behavior should learn a compact representation capable of accurately reconstructing commonly observed patterns. Observations that deviate substantially from these learned patterns are expected to generate larger reconstruction errors and can therefore be identified as anomalous.

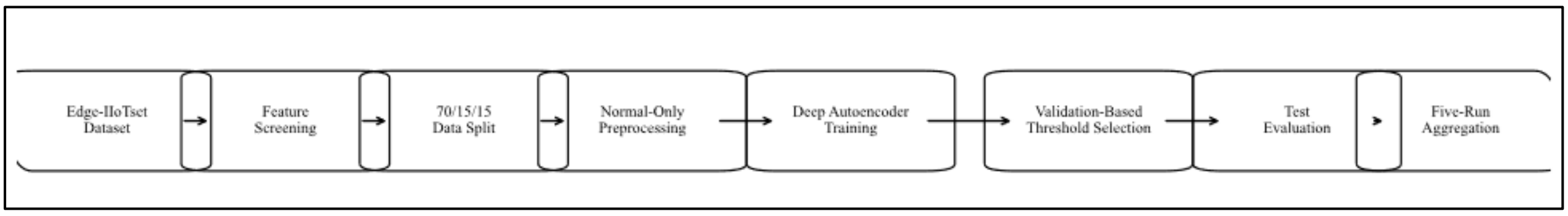

**Figure 1: Data preprocessing and anomaly-detection workflow.**

# 4 PROPOSED DEEP LEARNING MODEL

The proposed anomaly-detection model is a deep autoencoder designed to learn a compact representation of normal system behavior from the preprocessed Edge-IIoTset features. Rather than using attack labels during training, the model reconstructs normal observations and identifies potential anomalies from elevated reconstruction errors. This design allows the detector to model complex nonlinear relationships among system variables while remaining suitable for previously unseen or weakly represented attack patterns.

## 4.1 Deep Autoencoder Architecture

The model consists of a symmetric encoder-decoder architecture with **38 input features**. The encoder progressively compresses the input through hidden layers containing **24, 12, and 6 neurons**, while the decoder reconstructs the original feature representation using **12 and 24 neurons**, followed by a 38-neuron output layer:

$$38 \rightarrow 24 \rightarrow 12 \rightarrow 6 \rightarrow 12 \rightarrow 24 \rightarrow 38$$

Rectified Linear Unit (ReLU) activation functions were applied to the hidden layers to model nonlinear feature relationships, while the output layer used a linear activation function because the standardized input features could contain both positive and negative values. The six-neuron bottleneck serves as the compressed latent representation through which the model learns the dominant patterns associated with normal system activity.

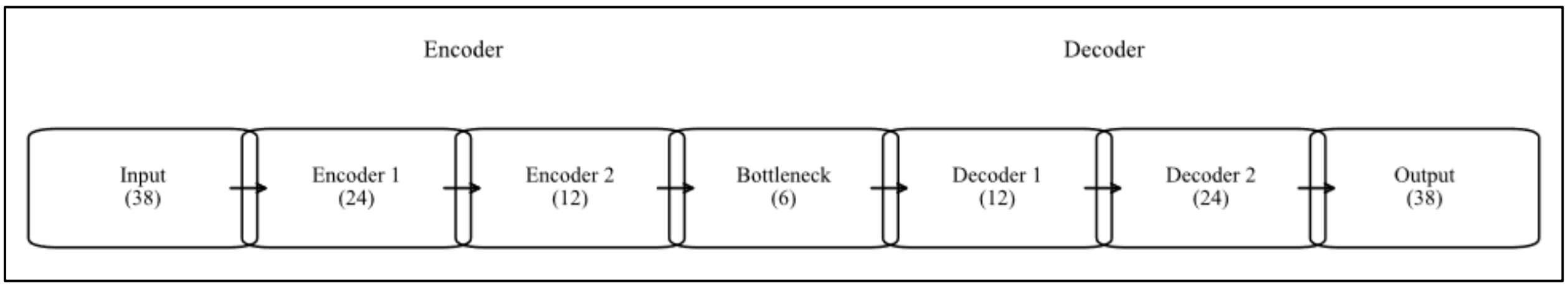

**Figure 2: Proposed Deep Autoencoder architecture.**

## 4.2 Reconstruction Error

The autoencoder was trained to minimize the mean squared reconstruction error between each normal input vector and its reconstructed output. For an observation $x_i$ with $p$ features, the reconstruction error was computed as:

$$RE_i = \frac{1}{p}\sum_{j=1}^{p}(x_{ij} - \hat{x}_{ij})^2,$$

where $X_{ij}$ denotes the original value of feature $j$, and $\hat{x}_{ij}$ represents the corresponding reconstructed value. Normal observations are expected to produce relatively small reconstruction errors, while anomalous observations that differ from the learned normal patterns generally produce larger errors.

## 4.3 Anomaly Threshold Selection

The anomaly decision threshold was determined exclusively from the validation dataset rather than the test set. Reconstruction errors were calculated for all validation observations, and precision and recall were evaluated across candidate thresholds. The threshold that maximized the validation F1-score was selected for each independent training run. An observation was subsequently classified as anomalous when

$$RE_i > T^*,$$

where $T^*$ denotes the validation-optimized reconstruction-error threshold. This procedure avoids selecting the operating threshold from the final test results and provides a balanced trade-off between anomaly detection and false alarms.

## 4.4 Model Training

The model was optimized using the Adam optimizer with an initial learning rate of 0.001 and mean squared error as the training loss. Training was performed for a maximum of 75 epochs using a batch size of 512. Early stopping with a patience of seven epochs was applied to limit overfitting, while the learning rate was reduced by a factor of 0.5 when validation loss failed to improve for three consecutive epochs. The best model weights observed during training were restored before evaluation.

## 5 EXPERIMENTAL DESIGN

The experimental design was structured to evaluate both the predictive performance and stability of the proposed Deep Autoencoder under repeated training conditions. All experiments used the same training, validation, and test partitions to ensure fair comparison across models. The proposed model was evaluated against a simpler autoencoder baseline, while performance was measured over five independent runs using different random seeds. Threshold selection was performed separately within each run using the validation set, and final performance was reported as the mean and standard deviation across runs.

### 5.1 Baseline Autoencoder

A simple autoencoder was implemented as the baseline model to determine whether the proposed architecture provided measurable performance gains. The baseline consisted of a single hidden bottleneck layer:

$$38 \rightarrow 12 \rightarrow 38$$

The baseline used the same preprocessing procedure, training data, optimizer, loss function, validation-based threshold selection, maximum number of epochs, and batch size as the proposed Deep Autoencoder. This ensured that differences in performance were attributable primarily to model architecture rather than experimental conditions.

### 5.2 Repeated Experimental Runs

To reduce dependence on a single random initialization, each model was trained independently across five runs using random seeds of **42, 52, 62, 72, and 82**. For every run, a new model was initialized and trained from scratch. The anomaly threshold was recalculated independently from the validation reconstruction errors so that each model was evaluated using its own validation-derived operating point.

Final performance was summarized using the arithmetic mean and sample standard deviation across the five runs:

$$\bar{x} = \frac{1}{n}\sum_{i=1}^{n} x_i$$

and

$$s = \sqrt{\frac{1}{n-1}\sum_{i=1}^{n}(x_i - \bar{x})^2},$$

where $x_i$ represents the metric obtained from run $i$, is the mean across runs, and $n$=5.

### 5.3 Evaluation Metrics

Model performance was evaluated using **accuracy, precision, recall, F1-score, ROC-AUC, PR-AUC, and false positive rate (FPR)**. Precision measures the proportion of predicted anomalies that were correctly identified, while recall measures the proportion of actual anomalies successfully detected. The F1-score provides a harmonic balance between precision and recall:

$$F1 = 2\frac{Precision \times Recall}{Precision + Recall}$$

The false positive rate was calculated as

$$FPR = \frac{FP}{FP + TN},$$

where $FP$ and $TN$ denote false positives and true negatives, respectively. ROC-AUC was used to assess overall ranking performance across classification thresholds, while PR-AUC was included because it provides a more informative assessment of anomaly detection when class distributions are imbalanced.

### 5.4 Model Comparison

The proposed Deep Autoencoder and baseline Autoencoder were compared using the mean performance obtained across the five repeated runs. Relative performance improvement was also calculated for the principal metrics to quantify the performance differences between the two architectures. For metrics in which higher values indicate better performance, relative improvement was computed as

$$Improvement(\%) = \frac{M_{deep} - M_{baseline}}{M_{baseline}} \times 100,$$

where $M_{deep}$ and $M_{baseline}$ denote the mean metric values of the proposed and baseline models, respectively. For false positive rate, improvement was expressed as a percentage reduction because lower values indicate better performance.

## 6 RESULTS

The proposed Deep Autoencoder was evaluated across five independent runs and compared with the Simple Autoencoder baseline using the same dataset partitions, preprocessing procedure, and validation-based threshold selection. Performance was assessed using accuracy, precision, recall, F1-score, ROC-AUC, PR-AUC, and false positive rate.

### 6.1 Deep Autoencoder Performance

Across the five runs, the Deep Autoencoder achieved an average accuracy of 0.9195 ± 0.0061, precision of 0.9602 ± 0.0204, recall of 0.7348 ± 0.0095, and F1-score of 0.8325 ± 0.0121. The mean false positive rate was 0.0114 ± 0.0060, indicating that the model maintained a relatively low rate of normal observations incorrectly classified as anomalous.

**Table 2: Deep Autoencoder performance across five independent runs**

| Metric | Mean ± SD |
|---|---|
| Accuracy | 0.9195 ± 0.0061 |
| Precision | 0.9602 ± 0.0204 |
| Recall | 0.7348 ± 0.0095 |
| F1-score | 0.8325 ± 0.0121 |
| ROC-AUC | 0.8825 ± 0.0339 |
| PR-AUC | 0.8640 ± 0.0286 |
| FPR | 0.0114 ± 0.0060 |

For visualization of model behavior, the run with random seed 72 was selected as the representative run because its F1-score (0.8338) was closest to the five-run mean F1-score (0.8325); Figures 3–7 are therefore based on this representative run.

The training and validation reconstruction losses showed consistent convergence, indicating that the model learned stable representations of normal system behavior. The log-scaled distribution of validation reconstruction errors further illustrates differences between normal and anomalous observations relative to the optimized anomaly decision threshold.

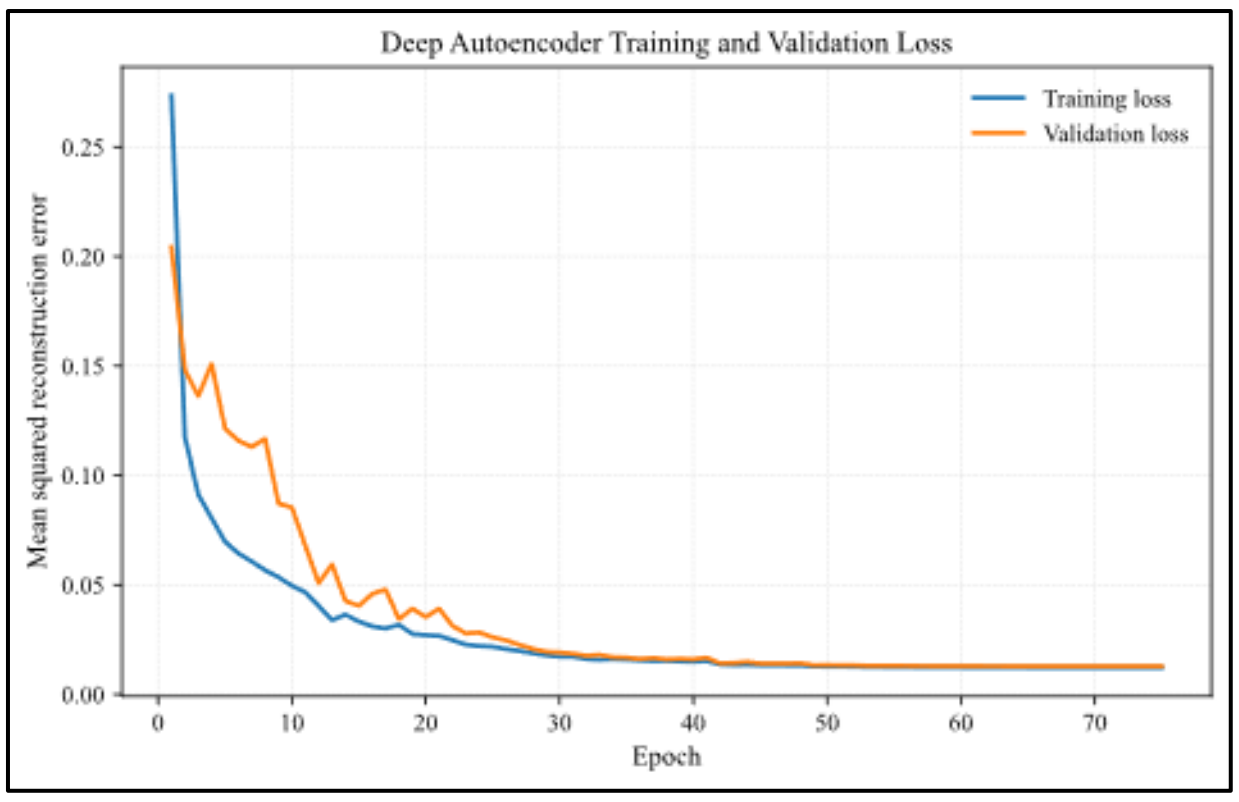

**Figure 3: Training and validation reconstruction loss.**

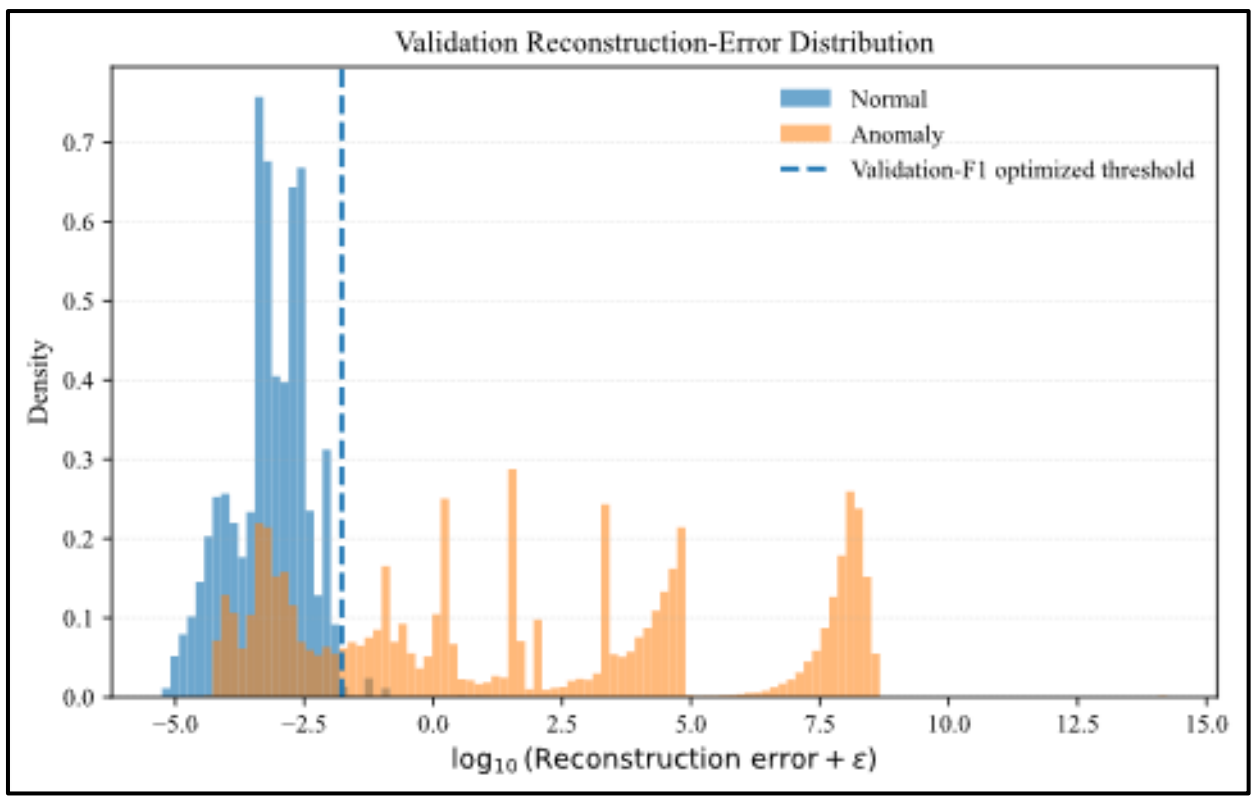

**Figure 4: Log-scaled validation reconstruction-error distribution and optimized anomaly threshold.**

The model obtained a mean ROC-AUC of 0.8825 ± 0.0339 and PR-AUC of 0.8640 ± 0.0286, indicating effective discrimination between normal and anomalous observations across decision thresholds. For the representative seed-72 run used for visualization, the corresponding ROC-AUC and PR-AUC values were 0.8636 and 0.8491, respectively.

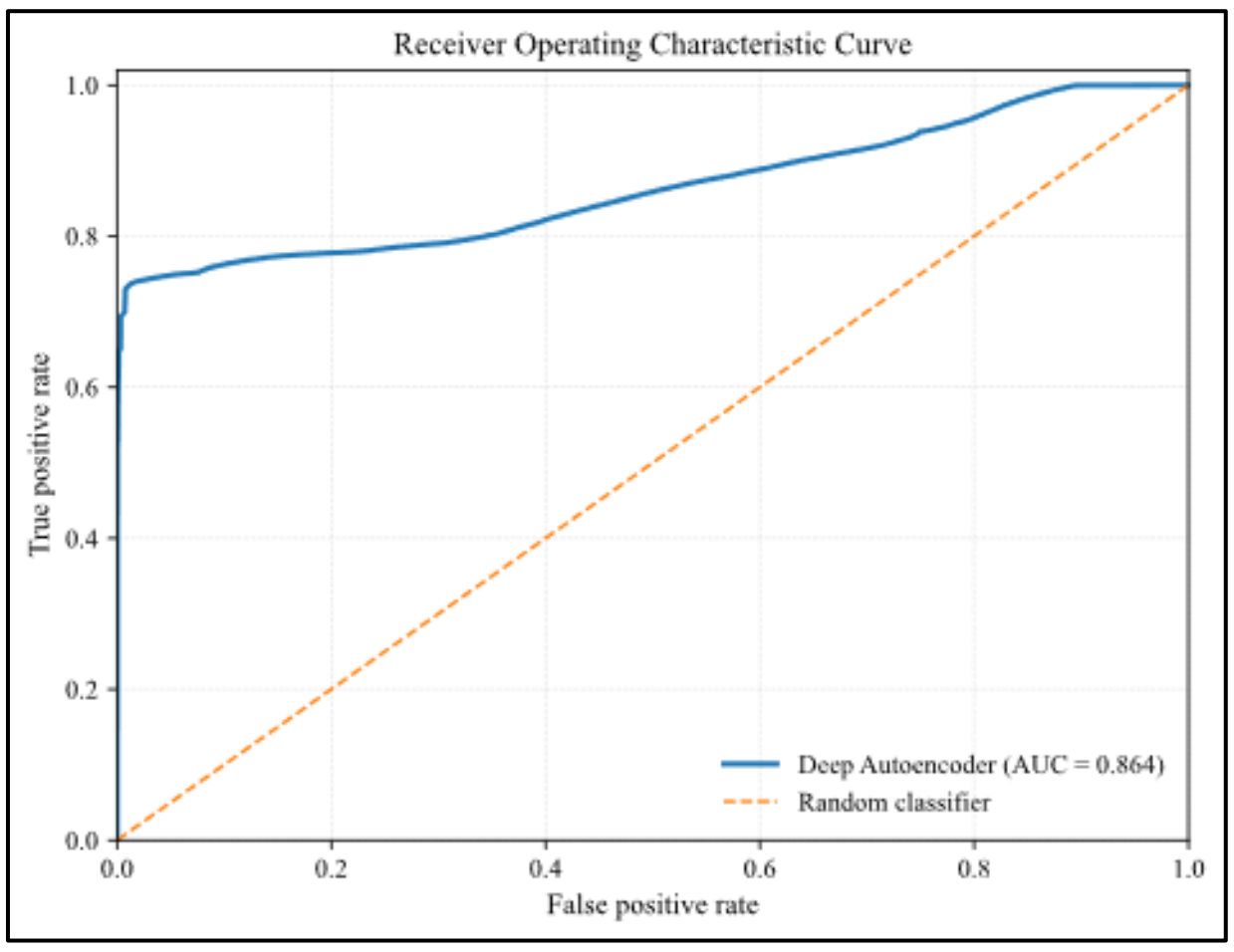

**Figure 5: Receiver operating characteristic curve.**

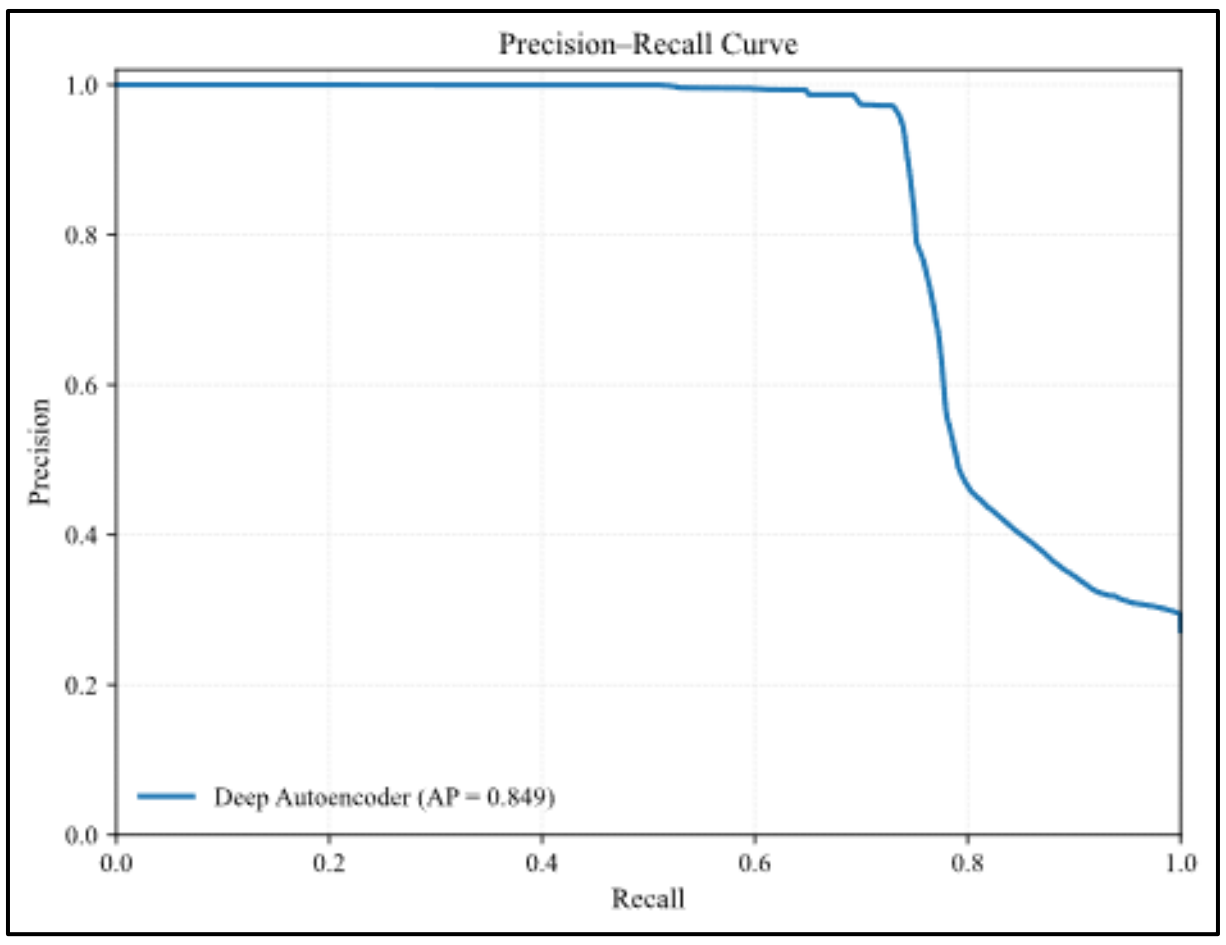

**Figure 6: Precision–recall performance of the Deep Autoencoder.**

The representative confusion matrix provides a detailed view of the model's correct detections and classification errors on the held-out test data.

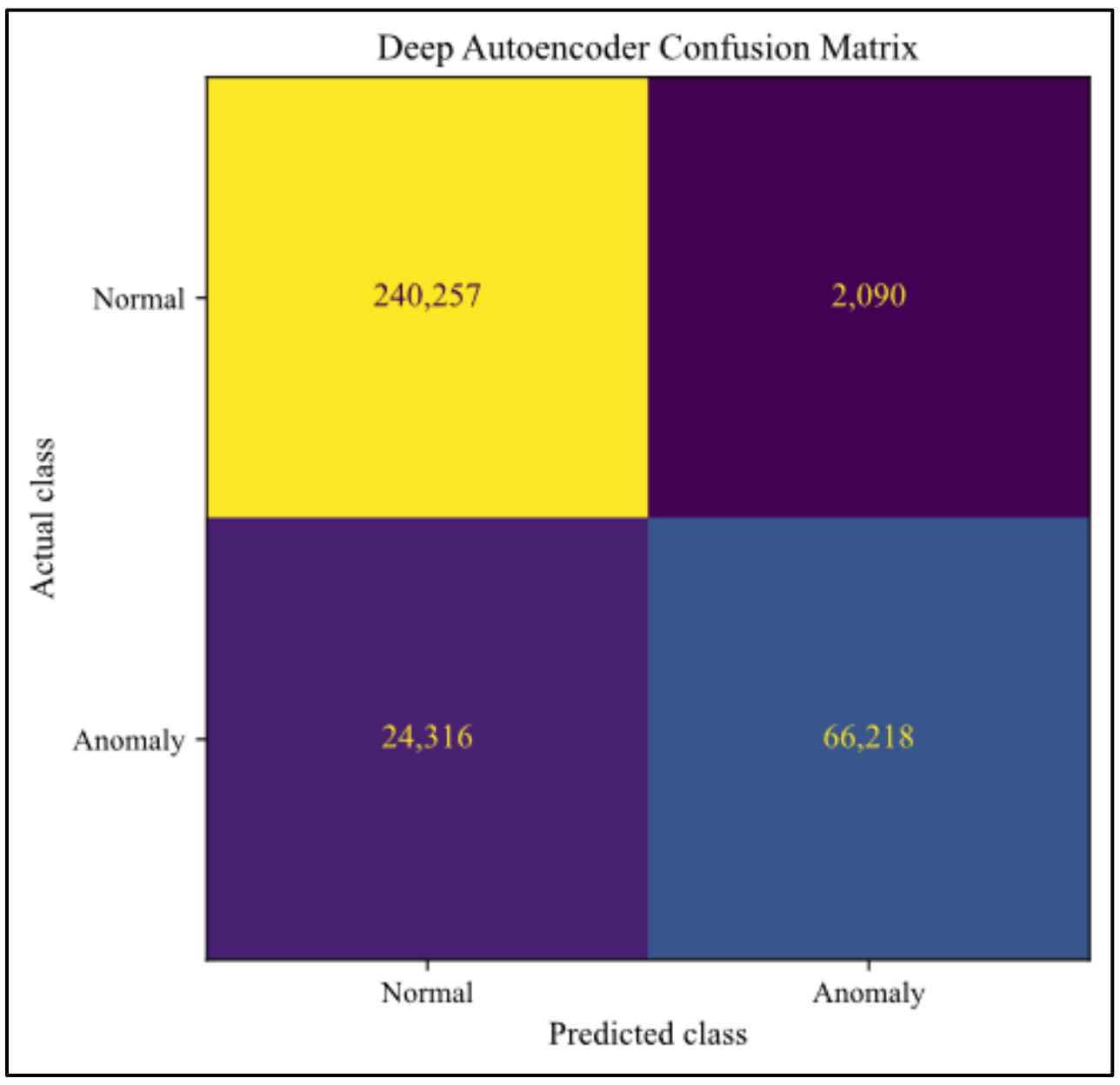

**Figure 7: Deep Autoencoder confusion matrix.**

The confusion matrix shows that the model correctly classified most normal and anomalous observations, while the remaining errors were concentrated primarily in false-negative cases.

The relatively small standard deviations observed across the five independent runs further indicate that the Deep Autoencoder maintained stable performance across different random initializations.

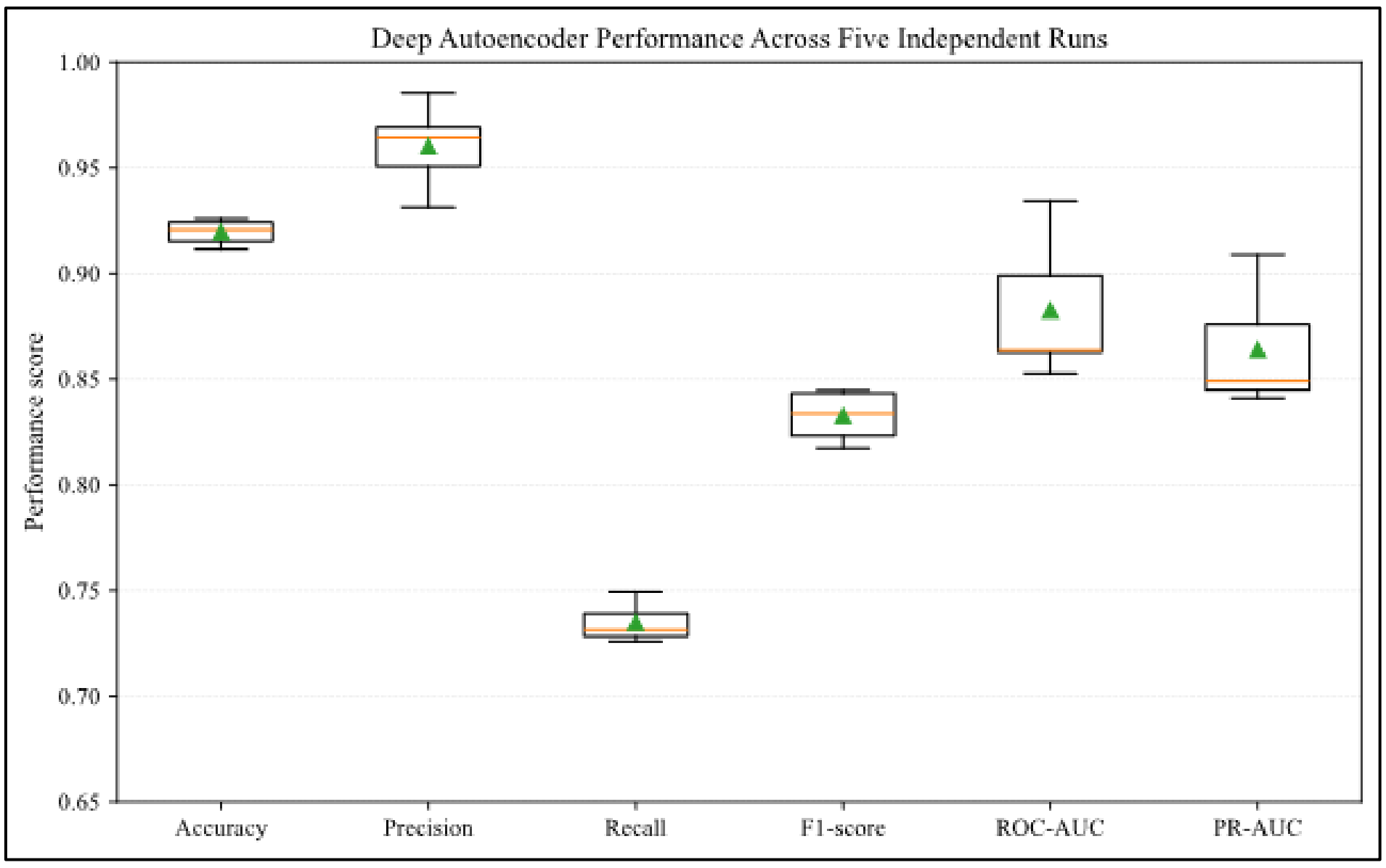

**Figure 8: Performance stability across five independent runs.**

Although some variation was observed across individual metrics, the overall performance remained consistent across repeated runs, supporting the stability of the proposed model.

## 6.2 Baseline Comparison

The Simple Autoencoder achieved lower average performance across most evaluation metrics. Its mean F1-score was **0.7310 ± 0.0143**, compared with **0.8325 ± 0.0121** for the Deep Autoencoder. Recall increased from **0.6017** to **0.7348**, while PR-AUC improved from **0.7983** to **0.8640**. The Deep Autoencoder also reduced the mean false positive rate from **0.0167** to **0.0114**.

**Table 3: Comparison of Simple and Deep Autoencoder models**

| Metric | Simple Autoencoder | Deep Autoencoder |
|---|---|---|
| Accuracy | 0.8795 ± 0.0071 | 0.9195 ± 0.0061 |
| Precision | 0.9315 ± 0.0318 | 0.9602 ± 0.0204 |
| Recall | 0.6017 ± 0.0125 | 0.7348 ± 0.0095 |
| F1-score | 0.7310 ± 0.0143 | 0.8325 ± 0.0121 |
| ROC-AUC | 0.8374 ± 0.0032 | 0.8825 ± 0.0339 |
| PR-AUC | 0.7983 ± 0.0038 | 0.8640 ± 0.0286 |
| FPR | 0.0167 ± 0.0079 | 0.0114 ± 0.0060 |

Relative to the baseline, the proposed model improved recall by approximately **22.1%**, F1-score by **13.9%**, PR-AUC by **8.2%,** and ROC-AUC by **5.4%.** The false positive rate was reduced by approximately **31.6%.**

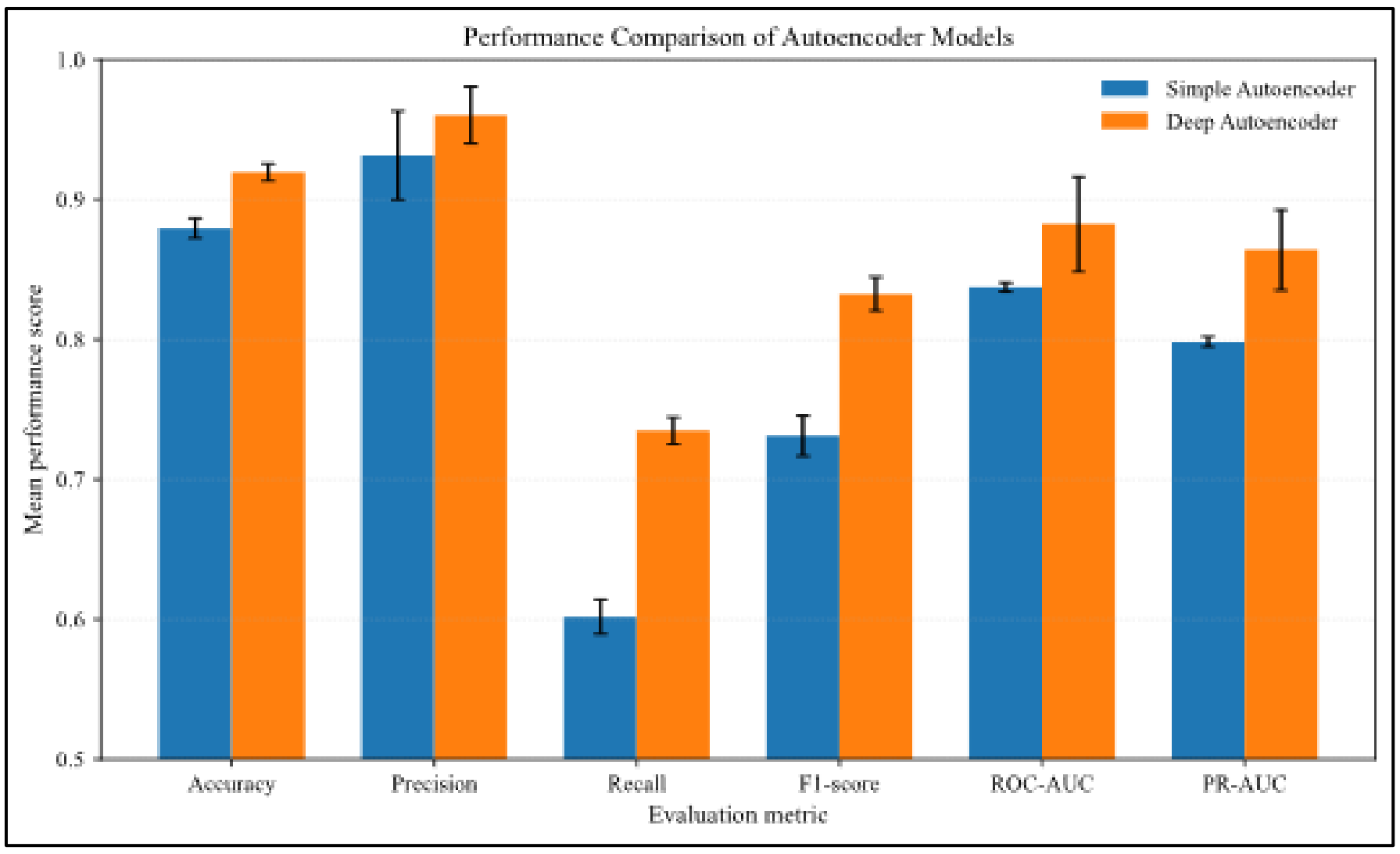

**Figure 9: Performance comparison of Simple and Deep Autoencoder models.**

The comparative results consistently favor the proposed architecture, particularly in recall and F1-score, indicating that its deeper and more compressed representation was associated with improved discrimination between anomalous behavior and normal system patterns.

### 6.3 Threshold and Detection Behavior

The anomaly threshold was determined separately for each run using the validation set. This procedure allowed the model to balance precision and recall without using the test data for threshold optimization. Threshold sensitivity analysis using the representative full-data run showed that excessively conservative thresholds produced very high precision but reduced anomaly recall. Validation-based F1 optimization provided a more balanced operating point and was therefore retained for the final experiments.

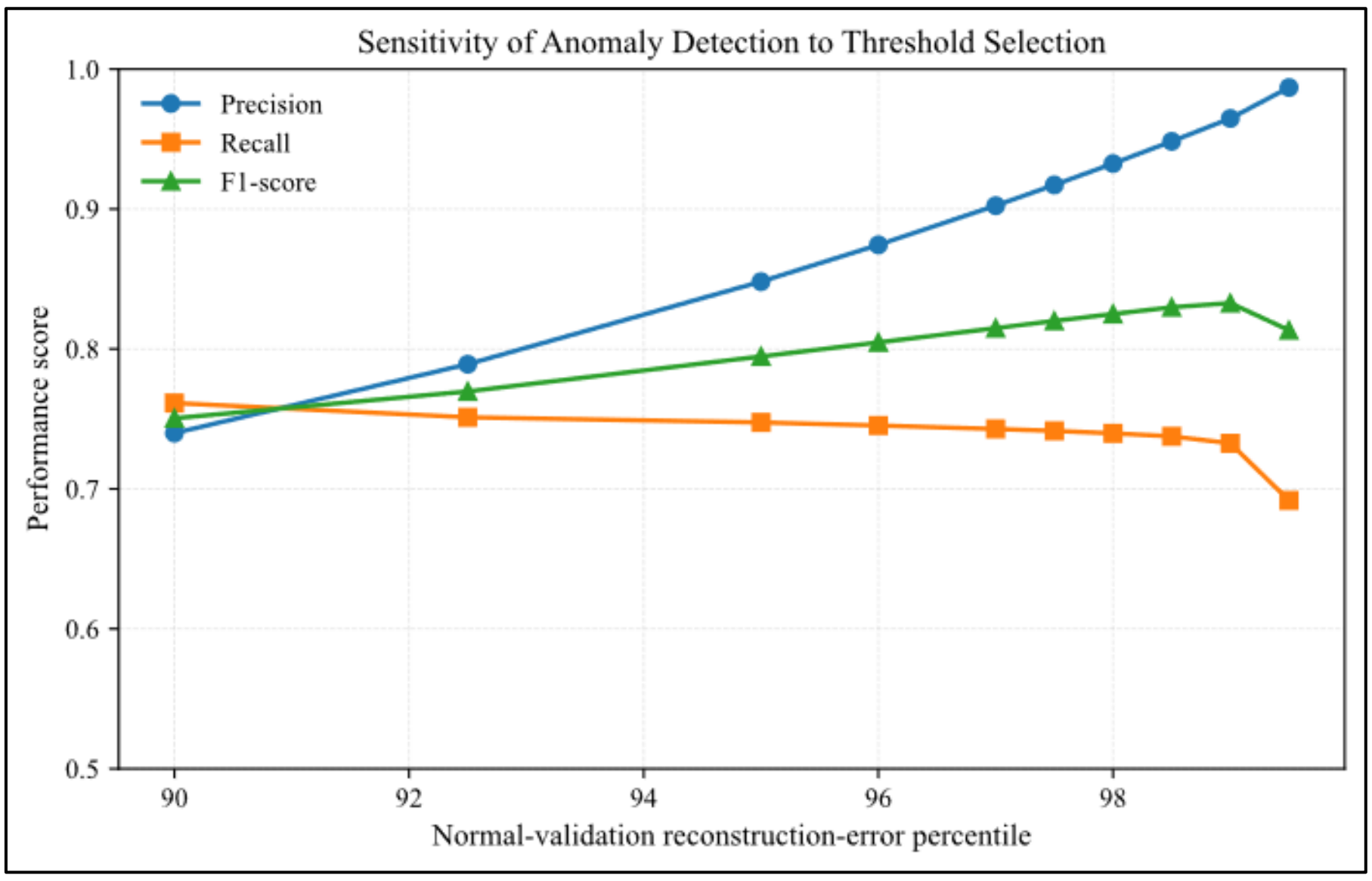

**Figure 10: Sensitivity of anomaly detection to threshold selection.**

The observed pattern illustrates the trade-off between precision and recall across candidate thresholds and supports the use of validation-based F1 optimization for the final anomaly decision rule.

## 7 DISCUSSION

The experimental results show that the proposed Deep Autoencoder provides a stronger and more balanced anomaly-detection capability than the Simple Autoencoder baseline. Across five independent runs, the proposed model achieved an average F1-score of 0.8325 ± 0.0121, compared with 0.7310 ± 0.0143 for the baseline. The improvement was particularly evident in recall, which increased from 0.6017 to 0.7348, indicating that the proposed model detected a substantially larger proportion of anomalous observations while maintaining high precision.

The performance gains suggest that the proposed deeper and more compressed encoder-decoder configuration captured nonlinear relationships among the retained system variables more effectively than the simpler baseline. By progressively compressing the 38-dimensional input into a six-dimensional latent representation, the model learned a more compact representation of normal behavior than the simpler 38–12–38 baseline. Consequently, anomalous observations produced reconstruction patterns that were more distinguishable from those associated with normal system activity.

The proposed model also maintained a relatively low mean false positive rate of 0.0114, compared with 0.0167 for the baseline. This result is important in anomaly-detection settings because excessive false alarms can reduce the operational usefulness of a detection system. At the same time, the validation-based thresholding procedure provided a more balanced trade-off between precision and recall than a fixed high-percentile threshold, demonstrating the importance of selecting the anomaly decision boundary independently of the test set.

The repeated-run analysis further showed that the model performance was generally stable across different random initializations. The relatively small standard deviations observed for accuracy, recall, F1-score, and false positive rate suggest that the reported results were not driven by a single favorable initialization. Some variability remained in ROC-AUC and PR-AUC, indicating that the ranking of reconstruction errors was somewhat more sensitive to stochastic model training than the final threshold-based classification performance.

Overall, the findings support the use of reconstruction-based deep learning for detecting abnormal behavior in cyber-physical environments. Rather than requiring the model to learn individual attack categories during training, the proposed approach models normal behavior and identifies deviations from that learned representation. This characteristic may be particularly useful in environments where anomalous events are diverse, evolve over time, or are incompletely represented during model development.

### *LIMITATIONS*

Although the proposed Deep Autoencoder achieved strong and stable anomaly-detection performance, several limitations should be acknowledged. First, the evaluation was conducted using a single publicly available dataset, and performance may differ in other cyber-physical environments with different devices, network conditions, or attack characteristics. Second, the study focused primarily on numerical features and excluded categorical and textual attributes, which may contain additional information relevant to anomaly detection. Third, the model was evaluated offline using historical observations rather than in a live streaming environment, where concept drift, latency, and changing system behavior may affect performance. Finally, the reconstruction-based approach identifies anomalous behavior but does not directly determine the specific attack type or root cause of an abnormal event. In addition, the stratified random partitioning strategy evaluates generalization across randomly separated observations rather than across temporally or operationally isolated sessions. Although this design provides a controlled comparison between models, future studies should examine temporal, device-level, or session-based partitioning to assess performance under stricter distribution shifts. Future work could therefore evaluate the model across additional datasets, incorporate temporal or categorical features, and investigate real-time deployment and attack characterization.

## 8 CONCLUSION

This study developed and evaluated a Deep Autoencoder for detecting anomalous behavior in cyber-physical systems using the Edge-IIoTset dataset. The proposed architecture learned normal system behavior through a compressed latent representation and identified anomalies using validation-optimized reconstruction-error thresholds. Across five independent runs, the model achieved an average accuracy of **0.9195 ± 0.0061**, precision of **0.9602 ± 0.0204**, recall of **0.7348 ± 0.0095**, F1-score of **0.8325 ± 0.0121**, ROC-AUC of **0.8825 ± 0.0339**, and PR-AUC of **0.8640 ± 0.0286**. Compared with the Simple Autoencoder baseline, the proposed model produced higher recall, F1-score, ROC-AUC, and PR-AUC while reducing the false positive rate. These findings demonstrate that the proposed reconstruction-based learning approach can provide an effective and reproducible method for identifying abnormal behavior in complex cyber-physical environments.

## STATEMENTS ON COMPLIANCE WITH ETHICAL STANDARDS

### *Acknowledgments*

I would like to express my appreciation to the Department of Mathematics & Statistics, College of STEM, and College of Graduate Studies at Austin Peay State University, Tennessee, for their support. I would also like to thank my family, friends, and colleagues for their encouragement and support.

### *Disclosure of conflict of interest*

The author declares no conflict of interest regarding this manuscript.